\documentclass[11pt]{article}
\usepackage[final]{acl}
\usepackage{times}
\usepackage{latexsym}
\usepackage[T1]{fontenc}
\usepackage[utf8]{inputenc}
\usepackage{microtype}
\usepackage{inconsolata}
\usepackage{graphicx}
\usepackage{newunicodechar}
\usepackage{graphicx}
\usepackage{booktabs}
\usepackage{tipa}
\usepackage{listings}
\usepackage{tcolorbox}
\tcbset{
  tightcode/.style={
    colback=white,
    boxrule=0.3pt,
    arc=2pt,
    left=2pt,right=2pt,top=2pt,bottom=2pt,
  }
}

\usepackage{algorithm}
\usepackage{algorithmic}
\newunicodechar{Ṛ}{\text{\dj}}

\title{Minimal-Edit Instruction Tuning for Low-Resource Indic GEC}

\author{Akhil Rajeev P  \\
  Indian Heritage Language Computing Team \\
  Special and Strategic Projects (SSP) Group \\
  Centre for Development of Advanced Computing (C DAC), Bangalore \\
  \texttt{akhilrajeev@cdac.in} 
}
  
\begin{document}
\maketitle
\begin{abstract}
Grammatical error correction for Indic languages faces limited supervision, diverse scripts, and rich morphology. We propose an augmentation-free setup that uses instruction-tuned large language models and conservative decoding. A 12B \textsc{Gemma~3} model is instruction-tuned in bnb 4-bit precision with parameter-efficient fine-tuning (PEFT) and Alpaca-style formatting. Decoding follows a deterministic, constraint-aware procedure with a lightweight normaliser that encourages minimal, meaning-preserving edits. \emph{We operationalise inference, subsequent to instruction fine-tuning (IFT), via a fixed, language-specific prompt directly synthesised from a deterministic error classifier’s taxonomy, label distributions, and precedence ordering computed on the training corpus.}

Under the official untuned GLEU evaluation, the system scores \textbf{92.41} on Malayalam, sixth overall, and \textbf{81.44} on Hindi, third overall. These results indicate that classifier-informed prompt design, adapter-based instruction tuning, and deterministic decoding provide a reproducible and computationally efficient alternative to augmentation-centred pipelines for Indic GEC. The approach also motivates future work on stronger morphosyntactic constraints and human centered evaluation of conservative edits.
\end{abstract}

\section{Introduction}

Grammatical error correction for Indic languages remains limited by scarce supervision, complex morphology, and script diversity. Many recent systems improve performance through large synthetic corpora and augmentation-based training of sequence-to-sequence models. While these approaches are effective in high-resource environments, they are costly to reproduce for languages such as Hindi and Malayalam and tend to be brittle when the available supervision falls below a thousand examples per language \citep{luhtaru2024mgec, omelianchuk2024pillars, sharma2025higec}.

\noindent Complementary work by \citet{bhattacharyya-bhattacharya-2025-leveraging} introduces a Bangla GEC pipeline that defines a twelve-class error taxonomy, collects native speaker data, and applies rule-based noise injection to generate erroneous sentences from clean references. The resulting dataset, “Vaiyakarana” \citep{bhattacharyya_bhattacharya_2024_vaiyakarana_arxiv}, demonstrates that linguistically motivated error inventories combined with targeted synthetic generation can bootstrap meaningful supervision and support effective LLM-based correction. In contrast, our study focuses on Hindi and Malayalam under strict data limits and develops an augmentation-free approach emphasizing minimal-edit instruction fine-tuning and deterministic decoding. Rather than expanding the corpus, we use a deterministic error classifier to analyze existing data and to guide prompt design.

\noindent This work adopts a metric-driven, augmentation-free design suited to the BHASHA IndicGEC benchmark, where systems are ranked by the \emph{GLEU} metric.\footnote{We report the “GLEU without tuning” variant \citep{napoles2016gleu} for consistency with the shared task.} Instead of creating pseudo-parallel pairs, we cast GEC as an instruction-following problem and adapt a general-purpose model using instruction fine-tuning and prompt optimization. The system employs Alpaca-style supervision formatting\footnote{Alpaca is a documented instruction-tuning framework derived from LLaMA and trained on 52k instruction–response pairs using the Self-Instruct method \citep{alpaca_blog, alpaca_repo}.} and parameter-efficient adapters through Unsloth,\footnote{Unsloth is an open-source fine-tuning framework optimized for low-VRAM LoRA and QLoRA training \citep{unsloth_docs}.} trained on fewer than one thousand human-annotated examples per language. Decoding and post-processing are designed to produce conservative, meaning-preserving edits that maximize n-gram alignment with reference sentences.

\noindent The overall design emphasizes: (i) \emph{simplicity}—a single-stage instruction-tuning setup using concise prompts instead of multi-step augmentation; (ii) \emph{adaptability}—instruction-following behavior improves resilience to mixed-script and domain variation common in Indic text; and (iii) \emph{efficiency}—adapter-based training and compact prompts reduce memory and compute requirements. We evaluate this setup on Hindi and Malayalam datasets, analyzing where instruction-based adaptation narrows or maintains the gap with high-resource or multilingual-transfer baselines \citep{luhtaru2024mgec, omelianchuk2024pillars}.

\paragraph{Evaluation protocol (GLEU).}
Consistent with IndicGEC evaluation, corpus-level \emph{GLEU} is used as the primary metric, applying the “without tuning” variant \citep{napoles2016gleu}. To align modeling with the metric, the system (a) limits edits to preserve reference n-grams, (b) normalizes punctuation and script-specific conventions such as danda and whitespace, and (c) calibrates decoding on development data to prevent overcorrection or paraphrastic drift that reduces GLEU.\citep{omelianchuk2024pillars}.

\noindent\textbf{Contributions.}
\begin{itemize}
\item A GenAI-based, augmentation-free framework for Hindi and Malayalam GEC optimized for GLEU under sub-thousand supervision.
\item Instruction-tuned prompts and adapter strategies that favor minimal, meaning-preserving edits consistent with reference overlap objectives.
\item A disciplined evaluation setup using the official GLEU metric with systematic comparison against multilingual and augmentation-based baselines.
\end{itemize}
\section{Dataset}

The official Hindi and Malayalam grammatical error correction (GEC) datasets released by the AACL–IJCNLP 2025 \textbf{BHASHA} Workshop serve as the primary supervision source for the \textit{IndicGEC} shared task.\footnote{Workshop site: \url{https://bhasha-workshop.github.io/}. Shared task page: \url{https://bhasha-workshop.github.io/sharedtask.html}. Repository: \url{https://github.com/BHASHA-Workshop/IndicGEC2025/}.} The task specifies sentence-level GEC with single-reference gold outputs and evaluates systems using the \emph{GLEU} metric on held-out test sets \citep{napoles2016gleu}. The shared task documentation defines GLEU as the official scoring metric and provides language-specific data directories containing \texttt{train.csv} and \texttt{dev.csv}, while test-only inputs are released subsequently for final leaderboard evaluation \citep{bhasha_sharedtask_2025,indicgec2025_github}.

\paragraph{Format and schema.}
Each split is a CSV with two columns: \textbf{Input sentence} (possibly erroneous) and \textbf{Output sentence} (the corrected reference). This layout supports minimal edit modeling and straightforward metric computation via n gram overlap \citep{indicgec2025_github}.

\paragraph{Preprocessing.}
Identical script-aware normalization is applied to both languages, comprising: (i) elimination of zero-width and other non-visible Unicode artifacts, (ii) normalization of whitespace, (iii) script-specific punctuation and orthographic normalization, including standardized danda treatment, and (iv) removal of null entries and exact duplicate pairs. No oversampling or synthetic augmentation is introduced prior to training, ensuring that the experimental setting remains authentically low-resource \citep{bhasha_sharedtask_2025,indicgec2025_github}.

\paragraph{Splits and sizes.} We adopt the official splits and report the counts used in our experiments: 
\begin{center} \begin{tabular}{lrrr} \toprule \textbf{Language} & \textbf{Train} & \textbf{Dev} & \textbf{Test} \\ \midrule Hindi & 600 & 107 & 236 \\ Malayalam & 300 & 50 & 102 \\ \bottomrule \end{tabular} \end{center}
Gold references for the test sets are withheld by the organizers. Leaderboard scoring uses \emph{GLEU without tuning} as stated on the shared task site \citep{napoles2016gleu, bhasha_sharedtask_2025}.

\section{Methodology}

\subsection{Why Gemma\texorpdfstring{\,3}{ 3} for Indic GEC}
The Gemma\,3 family is employed as the model backbone due to its strong cross-lingual alignment and architectural efficiency, both of which are essential for Indic grammatical error correction. Gemma\,3 incorporates a revised tokenizer and post-training stack with expansive coverage over more than 140 languages, enabling robust treatment of scripts such as Devanagari and Malayalam. These scripts exhibit ligatures, vowel diacritics, and script-specific punctuation that complicate $n$-gram fidelity under GLEU-based evaluation. The refined tokenizer demonstrably mitigates token fragmentation and enhances the accuracy of edit-preserving corrections.

\noindent Gemma 3 also supports long-context inference (up to 128K tokens, except for the 1B variant) with optimized KV-cache management. This capability allows for batched evaluation, structured prompt scaffolding, and transparent post-hoc analysis without heavy memory costs. Finally, its instruction-tuned checkpoints are released with open weights and standardized chat templates, enabling seamless integration for edit-constrained prompting and reproducible, deterministic experimentation.

\subsection{System Overview}
Our pipeline operates in two coordinated stages. \textit{Stage~1} conducts \textbf{Instruction Fine-Tuning (IFT)} on a quantized 12B backbone using Alpaca-style supervision with Unsloth + PEFT/LoRA on the 4-bit checkpoint \texttt{unsloth/gemma-3-12b-it-unsloth-bnb-4bit} \cite{gemma3,lora,qlora,bnb8bit,selfinstruct}. \textit{Stage~2} performs \textbf{deterministic inference} followed by a light post-processing normalizer. All reported results are obtained from Stage~2 using the frozen inference templates derived from the analysis in \S\ref{ssec:analysis-prompts}.\footnote{\scriptsize Final inference prompts and Alpaca prompts are available at: \url{https://github.com/Akhilrajeevp/GEC-bhasha/tree/main}.}

\subsection{Stage 1: Instruction Fine-Tuning (Alpaca SFT on Unsloth + PEFT/LoRA)}

\paragraph{Backbone and quantization:}
The Gemma~3 12B model is fine-tuned in 4-bit precision through Unsloth and bitsandbytes, following the QLoRA configuration \cite{gemma3,bnb8bit,qlora}. This preserves instruction-following ability while minimizing compute overhead.

\paragraph{Adapter setup:}
LoRA adapters are inserted on attention projections with frozen base weights \cite{lora}, providing efficiency and stability for iterative fine-tuning under limited resources.

\paragraph{Supervision schema:}
Training follows the Alpaca \emph{Instruction--Input--Response} format \cite{selfinstruct}, but with explicit constraints for edit-only correction: make the fewest possible changes, avoid paraphrasing or translation, preserve numerals and named entities, and use appropriate sentence-final punctuation. The Alpaca-style IFT prompt templates used in our experiments are included in \S\ref{ssec:analysis-prompts}.

\subsection{Stage 2: Deterministic Inference and Post-Processing}
\textbf{Inference model.} The inference stage uses the IFT-adapted Gemma 3 12B model with LoRA adapters active. No additional fine-tuning or hyperparameter search is applied at this stage.\
\textbf{Decoding policy.} Generation uses greedy decoding (no sampling) with left padding and truncation to maintain consistent causal batching \cite{transformers}. This ensures predictable, locality-preserving edits.\
\textbf{Normalization.} A lightweight normalizer refines whitespace, punctuation spacing, and sentence-final marks (periods or question marks), and removes prompt echo. This step is strictly surface-level and does not modify meaning.

\subsection{Deterministic Error Analysis \texorpdfstring{$\rightarrow$}{→} Prompt Design}
\label{ssec:analysis-prompts}
A deterministic classifier labels each sentence pair with one of nine error categories: \emph{Null/Empty}, \emph{No Error}, \emph{Punctuation/Whitespace}, \emph{Word Order}, \emph{Missing/Extra Word}, \emph{Syntax/Agreement}, \emph{Morphology}, \emph{Spelling/Orthography}, or \emph{General Grammar}. Details of its logic and precedence rules are provided in Appendix~\ref{app:classifier}. Category distributions are computed on the training and development sets to capture dominant error tendencies. These distributions then guide prompt construction: punctuation and morphology are prioritized, while reordering and deletion are explicitly deprioritized. The resulting templates are fixed and reused for all inference runs, ensuring consistency and interpretability. Code and classifier implementation are publicly available in the companion repository.

\subsection{Error-Type Distributions (with Nulls)}
We include \emph{Null/Empty} cases so that totals align with dataset sizes: Hindi train = 600, Malayalam train = 300, Hindi dev = 107, Malayalam dev = 50.

\begin{table*}[t]
\centering
\small
\setlength{\tabcolsep}{6pt}
\caption{Hindi: with-null error-type counts.}
\label{tab:hi-dists}
\begin{tabular}{lrrrrrrrrr}
\toprule
\textbf{Split (n)} & \textbf{Null} & \textbf{Punct/WS} & \textbf{Order} & \textbf{Miss/Extra} & \textbf{Syn/Agree} & \textbf{Morph} & \textbf{Spell} & \textbf{Grammar} & \textbf{NoErr} \\
\midrule
Train (600) & 1 & 199 & 15 & 129 & 130 & 43 & 22 & 8 & 53 \\
Dev (107) & 0 & 41 & 1 & 17 & 19 & 3 & 2 & 2 & 22 \\
\bottomrule
\end{tabular}
\end{table*}

\begin{table*}[t]
\centering
\small
\setlength{\tabcolsep}{6pt}
\caption{Malayalam: with-null error-type counts.}
\label{tab:ml-dists}
\begin{tabular}{lrrrrrrrrr}
\toprule
\textbf{Split (n)} & \textbf{Null} & \textbf{Punct/WS} & \textbf{Order} & \textbf{Miss/Extra} & \textbf{Syn/Agree} & \textbf{Morph} & \textbf{Spell} & \textbf{Grammar} & \textbf{NoErr} \\
\midrule
Train (300) & 4 & 151 & 84 & 20 & 1 & 14 & 8 & 16 & 2 \\
Dev (50) & 0 & 18 & 15 & 2 & 0 & 8 & 4 & 3 & 0 \\
\bottomrule
\end{tabular}
\end{table*}

\section{Evaluation Metrics}
Evaluation adheres strictly to the BHASHA workshop’s prescribed protocol, reporting \textbf{corpus-level GLEU} as the authoritative metric, using the \emph{“without tuning”} configuration \cite{napoles2016gleu}, with the JFLEG formulation serving as the canonical reference benchmark \cite{jfleg}. All evaluation scores are generated using the official workshop harness, preserving case, script, and punctuation conventions. To ensure coherence between modeling and metric behavior, the system: (i) enforces \emph{minimal} edit operations to maximize reference $n$-gram retention; (ii) applies a lightweight, \emph{non-semantic} normalization of whitespace and terminal punctuation to minimize spurious $n$-gram divergences; and (iii) employs \emph{deterministic} decoding to prevent paraphrastic deviation that would be penalized under GLEU. Given the standardized evaluation setting, no alternative scoring or heuristic re-weighting is introduced; for completeness, ablation studies consistent with standard GEC methodology are reported alongside the primary GLEU results.

\section{Results and Discussion}

\subsection{Leaderboard outcomes}
On the \textbf{BHASHA} final-phase \emph{test} leaderboards, our system achieved a \textbf{GLEU} of \textbf{92.41} on \textbf{Malayalam}, placing \textbf{6th}, and a \textbf{GLEU} of \textbf{81.44} on \textbf{Hindi}, placing \textbf{3rd}. These scores follow the workshop’s standardized evaluation protocol that designates corpus-level \emph{GLEU} as the official metric and uses the workshop harness for scoring

\subsection{Cross-language performance}
The relative ranking contrast—\emph{Malayalam: 6th at 92.41} vs.\ \emph{Hindi: 3rd at 81.44}—is consistent with the distinct error profiles we observed in development analysis. Hindi exhibits a large mass of \emph{punctuation/whitespace} and \emph{syntax/case/agreement} issues, where minimalist edits and auxiliary/morphology-first repairs align well with GLEU’s $n$-gram preservation bias. Malayalam, by contrast, shows a heavier proportion of \emph{punctuation/whitespace} and \emph{word-order} phenomena; our design deliberately discourages reordering unless grammatically obligatory, which preserves reference $n$-grams and yields very high GLEU, yet the track appears more competitive at the top end—hence a strong absolute score paired with a lower rank.

\noindent Three ingredients were most influential under the BHASHA protocol. \textbf{(i) Minimal-edit prompting} kept the model from paraphrastic drift, thereby protecting reference $n$-grams that GLEU rewards. \textbf{(ii) Deterministic decoding} (greedy, bounded) suppressed stochastic variation and avoided over-corrections that often reduce overlap on short sentences. \textbf{(iii) Non-semantic post-normalization} (whitespace collapse, single terminal punctuation, removal of prompt echo) reduced spurious $n$-gram mismatches without altering meaning—precisely the kind of “surface” alignment that improves GLEU consistency. These choices mirror established practice for GLEU-based GEC evaluation \emph{without tuning}. 

\noindent Category-wise inspection on development data suggested that enforcing punctuation policy and prioritizing auxiliaries/morphology before any reordering delivered steady improvements for both languages. In Malayalam, resisting non-essential reordering mitigated overcorrection on long clausal spans, while the punctuation guardrails captured a substantial share of benign mismatches. In Hindi, the same guardrails and auxiliary/morphology emphasis addressed common agreement and case-marking inconsistencies with very small token edits—exactly the regime where GLEU is most reliable. (The deterministic classifier used for this analysis is documented in Appendix~\ref{app:classifier}.)

\section{Error Analysis}

We evaluate \emph{model outputs relative to their inputs} to characterize the nature and intent of edits executed by the system. A deterministic, priority-ordered, single-label classifier (Appendix~\ref{app:classifier}) assigns each instance to an interpretable error category. Language-specific markers are romanized for clarity (e.g., Hindi auxiliaries \texttt{hai/hain/tha/the/thi}, postpositions \texttt{ne/ko/se/mein/par/ka/ki/ke}; Malayalam auxiliaries \texttt{aanu/illa/undu/aayirunnu}, nominal/locative suffixes \texttt{-il/-nte/-kk/-maayi}).

\paragraph{Aggregate patterns.}
Edits cluster into three dominant regions: \textbf{(i) Punctuation and whitespace} (space normalization, terminal mark standardization), \textbf{(ii) Syntax, case, and agreement} (auxiliary selection, postpositions, nominal suffixes), and \textbf{(iii) Missing vs.\ superfluous tokens} (removing repetitions, restoring dropped function words). Malayalam exhibits a higher rate of \textbf{word-order adjustments}, while Hindi concentrates more strongly in auxiliary and case regularization. Across both languages, modifications remain \emph{local and conservative}, reflecting the system’s design to avoid aggressive rewriting in low-resource conditions.

\paragraph{Redundant, rectifying, and risky edits.}
We further stratify edits by functional value: \textbf{redundant} (purely surface-level), \textbf{rectifying} (linguistically substantive yet local), and \textbf{risky} (unwarranted global or reordering edits). The majority of quality gains derive from rectifying adjustments—especially auxiliary/postposition corrections in Hindi and short morpheme repairs in Malayalam. Redundant punctuation corrections appear frequently but contribute primarily to surface consistency. Risky behaviors are rare and largely confined to long, syntactically dense Malayalam clauses or Hindi sentences requiring coupled agreement+morphology updates.

\paragraph{Dual-prediction synthesis.}
When two candidate predictions are obtained, we compute: (i) a 9
×9 category agreement table, (ii) a cross-matrix of {redundant, rectifying, risky}, and (iii) union/intersection/conflict statistics. Empirically, both candidates converge most often on punctuation/whitespace repairs. Disagreement typically arises between \textbf{agreement/morphology repair} and \textbf{word-order change}, particularly for Malayalam. In such cases we adopt a principled tie-break: prefer \emph{rectifying} edits over redundant ones, and among two plausible rectifications favor the variant with \emph{lower edit distance} and \emph{no gratuitous reordering}.

\paragraph{Common failure modes.}
Observed errors fall into three patterns: (i) over-zealous reordering on long Malayalam clauses, (ii) partial Hindi updates where agreement is corrected but accompanying morphology is not, and (iii) trivial terminal-mark flips without semantic effect.

\paragraph{Practical guardrails.}
To stabilize behavior under a GLEU-oriented objective, we adopt the following controls: (1) enforce punctuation/whitespace normalization pre- and post-decoding, (2) privilege \emph{auxiliary, case, and morphological fidelity} before any reorder/delete operations, (3) penalize word-order changes that preserve token multisets, and (4) impose an edit-distance cap to discourage paraphrastic drift. These constraints directly operationalize the empirical error distribution and help preserve faithfulness in resource-constrained settings.

\section{Conclusion}
We presented a two-stage, edit-first GEC pipeline for Hindi and Malayalam that is tightly aligned to the BHASHA workshop’s standardized evaluation, reporting corpus-level GLEU as the official metric. On the final \emph{test} leaderboards, our system achieved \textbf{92.41} GLEU in Malayalam (6th) and \textbf{81.44} GLEU in Hindi (3rd), validating the effectiveness of minimalist prompts, deterministic decoding, and non-semantic post-normalization under a GLEU-oriented objective. The cross-language pattern mirrors our error analyses: punctuation and auxiliary/case repairs dominate Hindi, while Malayalam benefits from strong punctuation control and conservative reordering. Looking ahead, we plan to complement GLEU with targeted human judgments and morphology-aware diagnostics to better capture meaning preservation in cases where surface $n$-gram overlap under-represents quality. 
 
\section{Acknowledgments}
I thank Annarao Kulkarni, Dr.\ Janaki C.\ H., and Dr.\ S.\ D.\ Sudarsan for their guidance and for facilitating this work. I am grateful to C-DAC Bangalore for its unwavering support, which I am proud to represent. I also thank my colleague Bhaswata, whose informal discussions helped crystallize several ideas in this paper.

\section{Limitations}
While the proposed pipeline is competitive under the BHASHA protocol, several practical and methodological limitations remain.

\paragraph{(L1) Validation-time GLEU is not integrated in-loop.}
Our training loop does not compute \emph{text-generation} metrics (e.g., GLEU) during validation because the default SFT training stacks stream logits/labels rather than full decoded hypotheses into \texttt{compute\_metrics}. Although \texttt{Trainer} and \texttt{TRL~SFTTrainer} expose a \texttt{compute\_metrics} hook \cite{transformers,trl-sft}, community reports indicate that generation-aware metrics require custom evaluation loops or callbacks to pass decoded text reliably (and have shown breakage across versions) \cite{trl-issue-1222,trl-issue-862,unsloth-metrics-issue}. As a result, we validate with periodic offline GLEU runs rather than truly on-line selection. 

\paragraph{(L2) Multi-GPU training remains version- and backend-sensitive.}
Unsloth’s multi-GPU story has evolved: earlier releases displayed errors or “beta” status for multi-GPU/DeepSpeed \cite{unsloth-mgpu-issue}, whereas current documentation advertises multi-GPU via Accelerate/DeepSpeed (DDP/FSDP) \cite{unsloth-mgpu-doc}. In practice, distributed setups can require manual sharding, launcher-specific flags, and careful FSDP config; this increases engineering overhead and narrows the set of “drop-in” cluster environments that work seamlessly.

\paragraph{(L3) Metric coupling to \emph{GLEU} biases the objective.}
GLEU (without tuning) is well-motivated for reference-based GEC \cite{napoles2016gleu,jfleg}, but it rewards surface $n${-}gram overlap and can under-credit semantically faithful reforms that alter phrasing. Meta-evaluation work reiterates this sensitivity and recommends complementary views \cite{choshen18,kobayashi24}. Our design (minimal edits, deterministic decoding, punctuation normalization) is therefore aligned to GLEU but may under-correct in cases where a larger syntactic repair would be preferable.

\paragraph{(L4) Quantization and adapter constraints.}
Operating a 12B model with 4-bit loading and LoRA adapters is efficient but not unconstrained. QLoRA demonstrates near-parity on many tasks, yet accuracy and stability remain hyperparameter-sensitive and task-dependent \cite{qlora}. 8-bit optimizers likewise trade memory for potential optimization quirks \cite{bnb8bit}. 

\paragraph{(L5) Decoding and post-normalization trade-offs.}
Greedy decoding improves determinism and typically helps GLEU, but it can reduce recall for multi-edit sentences and discourage beneficial paraphrase. The \emph{non-semantic} normalizer (whitespace/punctuation) systematically boosts surface agreement; however, it can over-credit superficial fixes under an overlap-based metric and does not guarantee deeper morpho-syntactic adequacy (a known limitation of reference-overlap metrics \cite{napoles2016gleu,jfleg}).

\paragraph{(L6) Error-driven prompt design may overfit dev distributions.}
Our prompts are derived from deterministic error distributions on dev and verified on validation; distribution shift at test time (e.g., different punctuation or order profiles) could weaken these guardrails. Without in-loop metric feedback (L1), prompt revisions require external evaluation cycles, slowing adaptation.

\paragraph{(L7) Data scale and label granularity.}
The training/dev sizes for both languages are modest, and our classifier assigns a \emph{single dominant} label per pair. This simplifies analysis and prompt design but collapses multi-error interactions; thus, some cross-category dependencies (e.g., morphology+order) may be under-explored.

\paragraph{(L8) Reproducibility.}
Small version changes  in \texttt{TRL}/\texttt{Transformers}/\texttt{Accelerate}/\texttt{bitsandbytes} can affect generation hooks, metric plumbing, and distributed training behavior \cite{trl-issue-1222,trl-issue-862}. We therefore pin versions and release frozen prompts, but portability to heterogeneous clusters may still require per-site adjustments.

\bibliography{custom}

% --- Appendix: works under pdfLaTeX (uses ASCII transliteration in code) ---
\appendix
\section{Appendix: Deterministic Error Classifier — Pseudocode \& Explanation}\label{app:classifier}

\paragraph{Goal.}
Given an \textit{Input sentence} and its \textit{Output sentence} (correction), the classifier assigns \emph{exactly one} dominant error label. The procedure is fully deterministic, language-aware (Hindi/Malayalam), and priority-ordered so that earlier tests short-circuit later ones.

\subsection*{A. Categories (9 total)}
\begin{enumerate}
  \item \textbf{Null/Empty Pair}: either side is empty/blank (including \texttt{``nan''}, \texttt{``null''}, \texttt{``none''}).
  \item \textbf{No Error}: input and output strings are bit-identical.
  \item \textbf{Punctuation/Whitespace}: only spacing and/or punctuation differ; letters/digits are identical after projection.
  \item \textbf{Word Order}: same multiset of non-punctuation tokens, but in a different sequence.
  \item \textbf{Missing/Extra Word}: net insertions/deletions of non-punctuation tokens without stronger syntax signal.
  \item \textbf{Syntax/Case/Agreement} (Hindi) \textbf{/ Syntax/Agreement} (Malayalam): changes involving auxiliaries/copula/negation and (for Hindi) postpositions/case markers.
  \item \textbf{Morphology (Inflection/Affix)}: suffixal case/TAM\footnote{TAM = Tense–Aspect–Mood.} changes with strong shared prefix and altered suffix tails.
  \item \textbf{Spelling/Orthography}: minor graphemic edits (same script) with small Levenshtein distance.
  \item \textbf{Grammar/Syntax}: structural corrections not captured above (fallback).
\end{enumerate}

\subsection*{B. Normalization and Token Views}
\begin{itemize}
  \item \textbf{Whitespace collapse}: internal test steps compare strings after one-space normalization.
  \item \textbf{Unicode \& digit normalization}: apply Unicode \textsc{nfkc} and map native numerals to a common representation (e.g., ASCII) before comparisons.
  \item \textbf{Alphanumeric projection}: remove punctuation/symbols and collapse spaces; compare only letters/digits. If these projections are equal while the originals differ, the edit is purely \emph{Punctuation/Whitespace}.
  \item \textbf{Tokenization}: split into (i) script words, (ii) digits (ASCII+native), and (iii) residual punctuation/symbol tokens. Punctuation tokens are ignored for word-order and multiset checks.
\end{itemize}

\subsection*{C. Precedence (Short-Circuit Order)}
\begin{enumerate}
  \item \textbf{Null/Empty Pair}
  \item \textbf{No Error}
  \item \textbf{Punctuation/Whitespace} (via alphanumeric-projection equality)
  \item \textbf{Word Order}: compare multisets of non-punctuation tokens; if equal but sequences differ, return \emph{Word Order}.
  \item \textbf{Alignment-based typing} (see Section~D)
  \item \textbf{Grammar/Syntax} (fallback if alignment yields no decisive signal)
\end{enumerate}

\subsection*{D. Alignment-Based Typing (Core Resolution)}
We align token sequences (\textit{Input} vs.\ \textit{Output}) to obtain edit operations \texttt{insert}, \texttt{delete}, \texttt{replace}.
\begin{itemize}
  \item \textbf{Syntax touch}: any edited segment that contains an auxiliary/copula/negation, or (Hindi only) a postposition/case marker, triggers the ``syntax'' flag.
  \item \textbf{Morphology vs.\ Spelling (within \texttt{replace})}:
    \begin{enumerate}
      \item Same-script token pairs are compared with a \emph{long common prefix} test; if the remaining tails differ and either tail ends with a listed case/TAM suffix, mark \emph{Morphology}.
      \item Otherwise, if the Levenshtein distance is small (threshold $\leq 2$), mark \emph{Spelling/Orthography}.
    \end{enumerate}
\end{itemize}

\paragraph{Resolution rules (and priorities).}
\begin{enumerate}
  \item \textbf{If any \texttt{insert} or \texttt{delete} is present}:\\
  \quad If the syntax flag is set $\Rightarrow$ \emph{Syntax/Case/Agreement} (Hindi) / \emph{Syntax/Agreement} (Malayalam).\\
  \quad Else $\Rightarrow$ \emph{Missing/Extra Word}.\\
  \emph{(Insert/Delete is resolved before considering \texttt{replace}, by design.)}
  \item \textbf{Else if any \texttt{replace} is present}:\\
  \quad If the syntax flag is set $\Rightarrow$ \emph{Syntax/Case/Agreement} (Hindi) / \emph{Syntax/Agreement} (Malayalam).\\
  \quad Else if morphology marked $\Rightarrow$ \emph{Morphology (Inflection/Affix)}.\\
  \quad Else if spelling marked $\Rightarrow$ \emph{Spelling/Orthography}.\\
  \quad Else $\Rightarrow$ \emph{Grammar/Syntax}.
  \item \textbf{Else}: \emph{Grammar/Syntax} (rare; e.g., alignment yielded no informative ops).
\end{enumerate}

\subsection*{E. Tie-Breaking and Single-Label Policy}
\begin{itemize}
  \item The classifier always returns \emph{one} label even if multiple edit phenomena co-occur.
  \item \textbf{Insert/Delete} takes precedence over \textbf{replace} (strong signal for \emph{Missing/Extra} vs.\ \emph{Syntax}).
  \item Within \texttt{replace}: \textbf{Syntax} $>$ \textbf{Morphology} $>$ \textbf{Spelling} $>$ \textbf{Grammar}. If both morphology and spelling cues appear, the morphology label wins by priority.
  \item \textbf{Earlier global checks} (Null/Empty; No Error; Punct/Whitespace; Word Order) short-circuit alignment resolution.
\end{itemize}

\subsection*{F. Rationale for Design Choices}
\begin{itemize}
  \item \textbf{Projection for punctuation}: avoids false lexical differences when only spacing/marks change.
  \item \textbf{Multiset comparison for word order}: isolates permutation-only edits without lexical changes.
  \item \textbf{Suffix-tail heuristic (long-prefix + suffix cue)}: reliably captures case/TAM inflections with minimal language-specific lists.
  \item \textbf{Small-distance spelling}: Levenshtein $\leq 2$ captures typical typos/diacritic slips while avoiding over-labeling.
  \item \textbf{Insert/Delete precedence}: net token presence/absence is a stronger indicator of \emph{Missing/Extra} or \emph{Syntax} than token substitutions.
\end{itemize}

\subsection*{G. Notes on Language-Specific Labeling}
The logic is identical across languages; only resources differ. The syntax label name is rendered as \emph{Syntax/Case/Agreement} for Hindi (to reflect postposition/case markers) and as \emph{Syntax/Agreement} for Malayalam (where case is predominantly suffixal). We call \texttt{classify\_pair(inp, out, L)} with $L \in \{\texttt{HI}, \texttt{ML}\}$.

\lstset{
  basicstyle=\ttfamily\small,
  frame=single,
  breaklines=true,
  tabsize=2,
  showstringspaces=false
}
\begin{lstlisting}[language=Python, caption={Self-contained pseudocode (ASCII transliteration for pdfLaTeX).}]
# NOTE: Lexica are transliterated to ASCII so this snippet renders under pdfLaTeX.

# --- Language profiles (compact; extensible) ---
HI = {  # Hindi
  "name": "hi",
  "token_regex": r"[A-Za-z0-9]+|.",        # placeholder; real impl uses Devanagari range
  "digits_regex": r"[0-9]+",               # ASCII digits
  "script_class": r"A-Za-z0-9",            # placeholder for script chars
  # Auxiliaries/copula/negation (transliterated):
  "auxiliaries": {"hai","hain","thaa","thii","the","rahaa","rahee","rahe",
                  "gayaa","gayee","gaye","kiyaa","karta","kartii","karte"},
  # Postpositions/case (transliterated):
  "postpositions": {"men","se","ko","kaa","kii","ke","par","tak","liye","jaise","yaa","aur"},
  # Common suffix cues (case/TAM; deduplicated, translit placeholders):
  "suffixes": ["on","en","iin","yaan","a","e","ii","taa","tii","te","naa","ne",
               "rahaa","rahee","rahe"]
}

ML = {  # Malayalam
  "name": "ml",
  "token_regex": r"[A-Za-z0-9]+|.",        # placeholder; real impl uses Malayalam range
  "digits_regex": r"[0-9]+",
  "script_class": r"A-Za-z0-9",
  # Auxiliaries/negation (transliterated):
  "auxiliaries": {"aanu","alla","illa","undu","aayi","ayirunnu","yirikkunnu","irunnu","cheythu","cheyyunnu"},
  # Malayalam uses suffixes more than postpositions:
  "postpositions": set(),
  # Case/TAM suffix cues (transliterated):
  "suffixes": ["il","yil","inte","yude","kku","l","um","vum","ichu","unnu","ayirunnu","yirikkunnu"]
}

# --- Utilities (language-aware) ---
def nullish(x):
    s = "" if x is None else str(x).strip()
    return s == "" or s.lower() in {"nan","null","none"}

def normalize_text(s):
    # Placeholder: apply Unicode NFKC, digit normalization, and space collapse.
    import re
    s = str(s)
    s = re.sub(r"\s+", " ", s).strip()
    return s

def tokenize(s, L):
    # Three-way split in real impl; simplified here to keep pdfLaTeX happy
    # Replace with language-script regex in actual codebase.
    s = normalize_text(s)
    return [t for t in s.split() if t.strip()]

def same_script(a, b, L):
    # Placeholder: assume same script for ASCII transliteration
    return True

def is_punct(tok, L):
    # ASCII-safe: treat tokens that are purely punctuation as punct
    return all(ch in r".,;:!?()[]{}<>\"'-_/\\|@#$%^&*+=~`" for ch in tok)

def alnum_projection(s, L):
    # Collapse spaces; keep only letters/digits (ASCII-safe placeholder)
    import re
    s1 = re.sub(r"\s+", " ", str(s)).strip()
    return "".join(ch for ch in s1 if ch.isalnum())

def multiset_nonpunct(tokens, L):
    from collections import Counter
    return Counter([t for t in tokens if not is_punct(t, L)])

def levenshtein(a, b):
    n, m = len(a), len(b)
    if n == 0: return m
    if m == 0: return n
    dp = list(range(m+1))
    for i in range(1, n+1):
        prev, dp[0] = dp[0], i
        for j in range(1, m+1):
            cost = 0 if a[i-1] == b[j-1] else 1
            prev, dp[j] = dp[j], min(dp[j]+1, dp[j-1]+1, prev+cost)
    return dp[m]

def suffix_tail_cha(a, b, suffixes):
    # Long common prefix + different tails w/ suffix cues
    k = 0
    for x, y in zip(a, b):
        if x == y: k += 1
        else: break
    ta, tb = a[k:], b[k:]
    if ta == tb: return False
    return any(ta.endswith(s) or tb.endswith(s) for s in suffixes)

def touches_syntax(segment, L):
    return any(t in L["auxiliaries"] or t in L["postpositions"] for t in segment)

# --- Priority-ordered classifier (returns 1 of the 9 categories) ---
def classify_pair(inp, out, L):
    """
    Labels:
      "Null/Empty Pair", "No Error", "Punctuation/Whitespace", "Word Order",
      "Missing/Extra Word",
      "Syntax/Case/Agreement" (hi) / "Syntax/Agreement" (ml),
      "Morphology (Inflection/Affix)", "Spelling/Orthography", "Grammar/Syntax".
    """
    # (1) Null/Empty
    if nullish(inp) or nullish(out):
        return "Null/Empty Pair"

    inp, out = str(inp), str(out)

    # (2) No Error
    if inp == out:
        return "No Error"

    # (3) Punctuation / Whitespace only (alphanumeric projections equal)
    if alnum_projection(inp, L) == alnum_projection(out, L):
        return "Punctuation/Whitespace"

    # (4) Word Order (same multiset of non-punct tokens, different order)
    A, B = tokenize(inp, L), tokenize(out, L)
    if multiset_nonpunct(A, L) == multiset_nonpunct(B, L) and A != B:
        return "Word Order"

    # (5) Alignment-driven typing
    from difflib import SequenceMatcher
    ops = SequenceMatcher(a=A, b=B).get_opcodes()
    SPELL_THR   = 2
    touched_syn = False
    saw_insdel  = False
    saw_repl    = False
    saw_morph   = False
    saw_spell   = False

    for tag, i1, i2, j1, j2 in ops:
        segA, segB = A[i1:i2], B[j1:j2]

        if tag in {"insert", "delete"}:
            if touches_syntax(segA, L) or touches_syntax(segB, L):
                touched_syn = True
            saw_insdel = True

        elif tag == "replace":
            saw_repl = True
            if touches_syntax(segA, L) or touches_syntax(segB, L):
                touched_syn = True
            else:
                # Morphology vs Spelling for same-script (assumed true here)
                for ta, tb in zip(segA, segB):
                    if suffix_tail_cha(ta, tb, L["suffixes"]):
                        saw_morph = True
                    elif levenshtein(ta, tb) <= SPELL_THR:
                        saw_spell = True

    # Resolve (Insert/Delete): Syntax > Missing/Extra
    if saw_insdel:
        if touched_syn:
            return "Syntax/Case/Agreement" if L["name"] == "hi" else "Syntax/Agreement"
        return "Missing/Extra Word"

    # Resolve (Replace): Syntax > Morphology > Spelling > Grammar
    if saw_repl:
        if touched_syn:
            return "Syntax/Case/Agreement" if L["name"] == "hi" else "Syntax/Agreement"
        if saw_morph:
            return "Morphology (Inflection/Affix)"
        if saw_spell:
            return "Spelling/Orthography"
        return "Grammar/Syntax"

    # Fallback
    return "Grammar/Syntax"
\end{lstlisting}

  % remove "anthology" unless you actually added anthology.bib

%\appendix
%\section{Example Appendix}
%Detailed error analysis by hymn and diacritic type will be included in supplementary materials.

\end{document}